\documentclass[letterpaper, 10 pt, conference]{ieeeconf}

\usepackage{graphicx} 
\usepackage{amsmath,bm}
\usepackage{amsfonts}
\usepackage{hyperref}
\usepackage{float}
\usepackage{algpseudocode}
\usepackage{multicol}
\usepackage{svg}
\usepackage{setspace}

\IEEEoverridecommandlockouts                              
\usepackage{support-caption}
\usepackage{subcaption}
\usepackage{xcolor}

\usepackage{algorithm}
\usepackage{mathtools, cuted}

\usepackage[noadjust]{cite}

\DeclareMathOperator*{\argmin}{arg min}
\DeclareMathOperator*{\argmax}{arg max}
\DeclareMathOperator*{\trace}{trace}
\usepackage{amssymb}  

\title{Constrained Stein Variational Gradient Descent \\
for Robot Perception, Planning, and Identification}

\author{Griffin Tabor$^{1}$ and Tucker Hermans$^{1,2}$
\thanks{This work was supported in part by the National Science Foundation under Grants 1846341, and 2149585.$^{1}$Griffin Tabor and Tucker Hermans are with the  University of Utah Robotics Center,
        SLC, UT, USA
        \{griffin.tabor,tucker.hermans\}@utah.edu%
$^{2}$Tucker Hermans is with NVIDIA, Seattle, WA, USA}
}

\begin{document}
\maketitle
\thispagestyle{empty}
\pagestyle{empty}

\begin{abstract}

Many core problems in robotics can be framed as constrained optimization problems. Often on these problems, the robotic system has uncertainty, or it would be advantageous to identify multiple high quality feasible solutions. 
To enable this, we present two novel frameworks for applying principles of constrained optimization to the new variational inference algorithm Stein variational gradient descent. Our general framework supports multiple types of constrained optimizers and can handle arbitrary constraints. We demonstrate on a variety of problems that we are able to learn to approximate distributions without violating constraints. Specifically, we show that we can build distributions of: robot motion plans that exactly avoid collisions, robot arm joint angles on the SE(3) manifold with exact table placement constraints, and object poses from point clouds with table placement constraints.
\end{abstract}

\section{Introduction}
Estimating uncertainty defines a core problem in robotics.
In robotic perception, where partial observability and noise are prominent, it is often desirable to reason about the world in a probabilistic way~\cite{prob-robotics}. Because robot motion and sensing is imperfect, there is an entire distribution of possible states of the system. Instead of trying to estimate the pose of a robot, for example, we approximate the distribution over possible poses. 
Modern robots might have uncertainty about their own pose, future states produced by a motion plan or poses and shapes of objects in the scene.
Similarly, an agent might compute the uncertainty over model parameters as part of system identification.
Then downstream tasks, like control or planning, can choose to make use the full distribution, or only take the single most likely value as a point estimate. With the entire distribution, one can find optimal control actions in expectation of the entire system properties distribution, as in \cite{tabor2023adaptive}. Alternatively, using state and motion uncertainty, you can use chance constraints like in \cite{chance-constraint}, where they solve for plans that have a less than 1\% chance of colliding with obstacles.

In many other robotics tasks, where an underlying distribution is less intuitive, it can still be useful to frame problems with respect to probability distributions.
For example, planning or control as inference, which aims to find optimal robot actions by framing the problem as a probability distribution and solve for a maximum a posteriori estimate.
It is often advantageous to generate multiple, diverse solutions. 
For example, in \cite{Qingkai}, the authors found that in their planning as inference problem, with a multi-modal distribution, different initializations sometimes resulted in different high quality object grasps. They argued that depending on the context, there were semantic reasons to choose between the multiple solutions.
This idea has led researchers to reframe some optimization problems to mirror the perception tasks and find a distribution of good solutions rather than a single solution~\cite{Thom-vi-mpc,vi_mp,toussaint2024nlpsamplingcombiningmcmc}. 

For all of the above scenarios, we can encode the problem as a target distribution $p(z|o)$, the likelihood of sampling $z$ conditioned on observation $o$.  
For all but the simplest of problems, directly performing inference on $p(z)$ is intractable. 
This motivates us to find an approximate probability distribution, that matches the target distribution in some way.
One approach, variational inference, selects a family of distributions $Q$ and finds the member $q^*(z)$ that closest matches the target distribution $p(z)$~\cite{Blei_2017}. Formally, this is expressed as 
\begin{equation}
    q^*(z) = \underset{q(z)\in Q}{\argmin}\quad{f(q(z),p(z))
    }\label{eq:variational}
\end{equation}
Where $f(q,p)$ is some statistical divergence between two distributions, often the Kullback-Leibler (KL) divergence \cite{kullback1951information,yi2022sliced,campbell2019universal}. Notable exceptions include \cite{WangVIMPC,NEURIPS2020_c928d86f}. 

Compared to traditional methods for variational inference, Stein variational gradient descent~(SVGD) provides a flexible particle based family of approximate distributions $Q$. In contrast to other particle based methods, like Markov chain Monte Carlo, SVGD is gradient based and parallelizable, enabling greater computational efficiency\cite{MCMCCHIB20013569,prob-robotics,MCMC_CVPR}.
Because of its flexibility and efficiency, SVGD has recently been applied to robotics for state estimation, system identification and planning~\cite{barcelos2021dual,lambert2021stein, honda2023stein,power2023constrained,tabor2023adaptive,sharma2023taskspace,lee2023stamp,stein-ICP-2106-03287,fan2021stein,stein_multi_robot}.

Traditional optimization based approaches to these problems often make use of constraints, collisions in motion planning\cite{altro,Schulman2013FindingLO}, object parameter feasibility in system identification\cite{bala_dynamics,inertia_consistent}, or bounding object motion between subsequent laser scans\cite{ICP_bounded_angle,ICP_Kinematic_constraints}. 
It is desirable that instead of only approximating a smooth cost function, we also guarantee each particle is feasible with respect to a set of constraints.
While there has been recent work in applying constrained optimization techniques to SVGD in specific scenarios, SVGD does not have a unified way of applying arbitrary constraints to arbitrary problems\cite{fabioLagrange,power2023constrained,zhang2022sampling}.

In this paper, we propose methodologies for applying algorithms for constrained optimization to the recent method of SVGD. 
To discuss our new formulations, first we will explain some background on, gradient descent, Stein variational gradient descent, and constrained optimization. Finally, we will use our understanding of variational inference and constrained optimization to propose two formulations for constrained optimization SVGD.
We show how the existing augmented Lagrangian method from \cite{fabioLagrange} fits into  a broader family of formulations that we call the $Q$ method. We name it thus, as we modify the variational family  $Q$ to only include distributions with particles that satisfy the constraints. 
We demonstrate how to construct other instantiations of the $Q$ method using alternative constraint formulations. 
Our second method, that we call the $p$ method, is based on approximating the target distribution $p$ that has no support in the infeasible regions. 

To help illustrate our methods, we demonstrate them on a simple 2D problem using: augmented Lagrangian, log barrier, and quadratic penalty methods.
For evaluation, we test our methods on multiple simple robotics problems and evaluate both the final results after the methods converge and the number of gradient steps required. Specifically, we demonstrate that both the $p$ and $Q$ methods can build a distribution of low acceleration collision free trajectories, that the $Q$ method can solve build a distribution of arm joint angles on the SE(3) Gaussian that meet placement constraints, and that the $Q$ method can build a distribution of feasible object poses based off of the iterative closest point algorithm.
We further show that our $Q$ method converges to the feasible set faster.  We observe this speed advantage gets more dramatic on equality constrained problems when the feasible set has fewer degrees of freedom.

\section{Related Work}
While SVGD is applicable to a wide range of problems, there are a few key areas of robotics that seem particularly applicable. There is related work in the areas of state estimation, system identification, planning, and other work on incorporating constraints into SVGD.
In \cite{stein-ICP-2106-03287} authors construct a distribution over transformation matrices where the likelihood of a transformation is how well it aligns two point clouds. However, there is inherent uncertainty and created by the noise and the symmetries of objects that make solving for a single true transformation impossible. SVGD lets them build a multimodal distribution over the possible transformations. Different symmetries result in different distributions of possible transformations. Other methods, like using a Kalman Filter\cite{efk_ICP} is built on a Gaussian assumption. 
Both \cite{fan2021stein, stein-particle-filter} propose using SVGD in place of a resampling step in a particle filter, a popular robotics state estimation algorithm that takes in a map of the environment and robot sensors to estimate robot pose.

In \cite{DBLP:journals/corr/abs-2103-12890} and \cite{tabor2023adaptive} the authors use SVGD to approximate distributions over system parameter conditioned on a dataset of robot motion. 
Producing a distribution of system parameters lets the robot reason about the dynamics in a probabilistic way, specifically picking controls that are optimal in expectation under their approximate dynamics distribution. A more traditional approach to building a distribution of system parameters would be a Kalman filter, which again has a Gaussian assumption and \cite{kf_state_parameters}.

Using SVGD to build a distribution of motion plans has been shown as a valuable way to sample low cost and thus high quality plans\cite{barcelos2021dual,lambert2021stein, honda2023stein,power2023constrained}.  Finding multiple unique high quality plans is helpful potentially for replanning, adversarial scenarios and as a global search strategy to find the best motion plan as typical optimizers converge only to local minima. This aligns well with other recent work using variational inference to build distributions of motion plans\cite{yu2024stochasticmotionplanninggaussian}.

In \cite{lambert2021stein} the authors use a soft constraint formulation, where trajectories that collide with objects have low but non-zero probability.
In \cite{fabioLagrange} the authors proposed using the gradient of the augmented Lagrangian objective function to move particles towards satisfying equality constraints.
In \cite{power2023constrained}, the authors recently proposed constrained Stein variational trajectory optimization a way to generate a diverse set of constraint-satisfying trajectories. 
Their method, which builds on \cite{zhang2022sampling} has two constraint satisfying mechanisms, constructing kernel functions for SVGD that avoid moving into violation and taking Gauss-Newton steps away from constraint violations.
They provide little theoretical basis for mixing a first order method like SVGD and a second order method like Gauss-Newton. In \cite{zhang2022sampling}, authors used a nonstandard, but still first order gradient-based method, to lower constraint violations. 

\section{Background}
Numerical optimization is a rich field in mathematics dedicated to solving equations of the form.
\begin{align}
  \underset{x}{\argmin} f(x) \notag
\end{align} 
There are numerous algorithms for solving these problems, but for the purpose of this discussion we will focus on gradient descent. 
Gradient descent computes the steepest descent direction of the function $f$, its negative gradient, and then makes a small step in that direction(Eq.~\ref{eq:gd}).
\begin{align}
\label{eq:gd}
  x_{k+1} = x_{k} - \epsilon_k \nabla_x f(x)
\end{align} 

\subsection{Stein Variational Gradient Descent}
Stein variational gradient descent (SVGD) uses a set of particles as an implicit distribution $q$ to approximate a target distribution $p$. It defines a one-to-one distribution transformation $T_\epsilon$ such that $T_\epsilon(z) = z + \epsilon \phi(z)$ \cite{liu2019stein}. This transform is a single step in an iterative algorithm. The goal then is to find a transform that will bring $q$ closer to $p$, and thus minimizes $\textit{KL}\left( q_{[T]} || p \right)$.
Note, by definition, a transformation on $q$ and an inverse transformation on $p$ has the same effect on KL divergence. 
\begin{equation}
    \textit{KL} \left( q_{[T]} || p \right) = \textit{KL} \left( q || p_{[T^{-1}]} \right)
\end{equation} 

By taking the gradient of KL divergence with respect to epsilon, we have a function for finding how transformation directions will affect KL divergence. 

\begin{equation}
   \nabla_\epsilon \textit{KL} \left( q || p_{[T^{-1}]} \right) = - \underset{z \sim q}{\mathbb{E}} \left[ 
   \begin{matrix}
   \nabla_\epsilon \log p(z + \phi(z))^\top \phi(z)  +   \\
     \trace \left( \left(I + \epsilon \phi(z) \right)^{-1} \cdot \nabla_z \phi(z) \right) 
    \end{matrix}
    \right]
\end{equation}

If we evaluate this gradient at $\epsilon = 0$  we get $- \underset{z \sim q}{\mathbb{E}}\left[ \nabla_\epsilon  
\log p(z)^\top \phi(z) + \trace \left( \nabla_z \phi(z) \right) \right] $. It follows if we search for $\phi$ in some functional family $\Phi$ that maximizes this quantity it will be the steepest descent direction, within $\Phi$, to improve the KL divergence. 
\begin{align}
    & \phi^*(x) = \underset{\mathbf{\phi}(z)\in \Phi}{\argmax} -\!\!\!{\underset{z\sim q}{\mathbb{E}}}\left[ \nabla_\epsilon  
\log p(z)^\top \phi(z) + \trace \left( \nabla_z \phi(z) \right) \right] 
\end{align}
Liu et al. \cite{liu2016kernelized} shows if we select the family $\Phi$ to be functions within the unit ball of the reproducing kernel Hilbert Space, for a kernel function $k(z)$, this optimization has a closed form solution.
\begin{equation}
    \phi^* = \underset{z \sim q}{\mathbb{E}} \left[k(z) \nabla_z \log p(z) + \nabla_z k(z)\right]
\end{equation}
SVGD approximates this expectation using a finite set of particles and updates them incrementally using the functional gradient $\phi^*$. This non-parametric form provides more capacity and flexibility than a fixed family of analytic distributions.
\begin{align}
    z_i^{l+1} &\leftarrow z_i^{l} + \epsilon \phi^*(z_i^{l}) \quad  \\
    \quad  \phi^*(z) &\approx \frac{1}{n} \sum_j^n \left[ k(z_j^l,z) \nabla_{z_j^l} \log \left( p \left(z_j^l \right) \right) + \nabla_{z_j^l} k(z_j^l,z) \right]  \label{eq:stein_update}
\end{align}
There is an intuitive explanation for the terms in the update equation, the first term is the weighted average of the gradients of nearby particles and the second term is a repulsive term pushing particles away from each other.

\subsection{Constrained optimization}
 Constrained optimization extends numerical optimization problems to include hard constraints over the decision variables. The problem becomes finding the lowest cost point within the feasible region, the set where all constraints are satisfied.
\begin{align}
  \argmin & f(x) \\
   h_i(x) = 0 & \quad \text{for all i}  \notag\
  \\  g_j(x) \leq 0  & \quad \text{for all j}   \notag
\end{align} 

There are many classes of algorithms that can solve constrained optimization functions\cite{NoceWrig06}. Here will we discuss: projection methods, Lagrange multipliers, penalty methods, and interior point methods. 

\par{Projection methods} apply an update $\hat{x}_{k+1} = x_{k} - \epsilon_k \nabla f(x)$ and then find the point $x_{k+1}$ that is closest to $\hat{x}_{k+1}$ and in the feasible set. Given an infeasible start, they get a feasible solution in 1 step and then stay in the feasible region. 
Traditionally used on linear constraints where the projection is easy to compute.
E.g.The projection of $\hat{x}_{k+1} =  (-1,2)$ onto constraint $x \leq 0$ is $(-1,0)$. 

\par{Lagrange multipliers} let us handle linear equality constraints when computing the Newton step as part of solving a linear set of equations. This is a common update in numerical optimization, e.g., second order methods use a second order Taylor expansion and solve the step direction to set the gradient to zero. When the linear equality constraints are included in the Taylor expansion, second order methods solve the equality constraints exactly. 

\par{Penalty methods} convert a constrained optimization problem to a sequence of unconstrained optimization problems with a penalty for violating the constraint. The quadratic penalty for equality constraints gives an unconstrained solver the following problem $\argmin  f(x) + c \sum_k h_k(x)^2$ and increases $c$ iteratively until the solution is in the feasible region. 
The cost function for an augmented Lagrangian formulation with arbitrary equality and inequality constraints is shown in Eq~\ref{eq:augla}~\cite{toussaint2014novel}.
\begin{align}%
\hat{f}(x) =& f(x) + \lambda^T h(x) + \gamma^T  g(x) + c \sum_i h_i(x)^2 \notag\\ 
&+ d \sum_j [g_j(x)>0 \lor  \gamma_j>0 ]g_j(x)^2 
\label{eq:augla}
\end{align}%

This general formulation has an inner unconstrained optimization for a set of fixed parameters and an outer parameter update.

\par{Interior-point} methods require a feasible initialization and then create a ``barrier'' keeping iterates from leaving the feasible region. Similar to penalty methods, they augment the cost function and use an unconstrained optimizer on the augmented cost function. The log barrier function for inequality constraints is shown below.
\begin{equation}
    \hat{f}(x) = f(x) - \mu \sum_j \log(-g_j(x))
\end{equation}
This new cost function in undefined when it violates a constraint and has high cost the closer $x$ gets to violating the constraints. For smaller and smaller values of $\mu$ the effect, it has within the feasible set is minimized while still maintaining a barrier. As $\mu$ approaches $0$ the barrier, term approaches the indicator function.
One approach to support infeasible initializations is the relaxed log barrier, which acts as an augmented Lagrangian when infeasible and a log barrier when feasible~\cite{relaxedlogbarrier,relaxedlogbarrier_original}. This relaxed version of the log barrier is shown below
\begin{align}
    \hat{f}(x) &= f(x) + \mu \sum_j D(g_j(x)) \\
    D(g) &= \left\{ \begin{array}{ll}
        -\log(-g) & g \leq -\delta \\
        \frac{1}{2}\left( \left(\frac{g + 2\delta}{\delta}\right)^2 -1 \right) - \log(\delta) & g_j(x) > -\delta
    \end{array} \right.
\end{align}

\section{Method}
One difficulty with SVGD for robotics tasks, is its lack of ability to handle constraints, collision constraints and actuator constraints are incredibly common in trajectory optimization for example. 
In these constrained tasks, given a cost function $f(x)$ we would like to approximate 
\begin{align}
     p(x) = \frac{1}{Z}    \left\{\begin{array}{ll}
      e^{-\alpha f(x)} & h_i(x)=0, \forall i \bigwedge \\
      & g_j(x)\leq 0, \forall j \bigwedge \underline{x} \leq x \leq \bar{x} \\
      0 & \text{otherwise} \label{eq:true_dist}
\end{array}  
\right. 
\end{align}
The cost function is only helpful for comparing the quality of values within the feasible set, outside the feasible set we want $0$ probability mass. This creates a probability distribution with poor gradient information for SVGD to optimize with.

Building upon Eq.~\ref{eq:variational}, we present two formulations to approximate $p$, incorporating the constraints by modifying $Q$ or modifying $p$.
\subsection{Modifying Q}
Remembering SVGD defines $Q$, the family of distributions we are searching over, as all unique positions of $n$ particles we can instead create a feasible subset $\hat{Q}$. This feasible subset is all unique positions of $n$ particles such that each particle is in the feasible region.%
\begin{align}
    \bm{q}^*(x) &= \underset{\bm{q}\in \hat{Q}}{\argmin}\quad KL(\bm{q},p) \label{eq:qstar}\\
    \hat{Q} &= \left\{ \bm{q}_i \mid  \left( h(q_k)=0 \And g(q_k)\geq0 \right) \forall k 
\right\} 
\end{align}%

The full $Q$ method algorithm is shown in Alg.~\ref{algo:Q}.
First, we construct an example $p(x) = \frac{1}{Z} e^{-\alpha f(x)}$,  Then we attempt to approximate the distribution $q$ that approximates $p$ while searching in the family $\hat{Q}$. 
Existing constrained optimization suggests an efficient way to solve Eq.~\ref{eq:qstar} is to iteratively solve ${\argmin}\quad KL(\bm{q},p) + L_\theta(\bm{q})$, given a soft constraint formulation $L$ parameterized by $\theta$.  Gradient descent applied to this problem is 
$\bm{q} = \bm{q} - \epsilon( \nabla_{\bm{q}} KL(\bm{q},p) + \nabla_{\bm{q}} L_\theta(\bm{q}))$. 

\begin{algorithm}
\caption{Q Constrained SVGD}
\label{algo:Q}
\begin{algorithmic}[1]
    \State $q \gets k$ points sampled from distribution $D$
    \State $p \gets  \frac{1}{Z} e^{-\alpha f(x)} $
    \State $b \gets \text{min}(\text{max}(\underline{x},x),\bar{x})$
    \While{not converged}
    \While{not converged}
    \State $\forall i \in k \quad \bm{q}_i \gets b\left(\bm{q}_i + \epsilon ( \phi^*_p(\bm{q}_i) - \nabla_{\bm{q}_i} L_\theta(\bm{q}_i) ) \right)$
    \EndWhile
    \State $\forall i \in k \quad \theta_i \gets \text{constraint\_param\_update}(\theta_i,\bm{q}_i)$
    \EndWhile

\end{algorithmic}
\end{algorithm}

Using $\phi^*$ as $-\nabla_{\bm{q}} KL(\bm{q},p)$, we select the function $L_\theta$ from the existing constrained optimization literature. After computing the desired gradient step, we use projection to efficiently and exactly keep each particle inside the box constraints.

This general formulation supports all existing soft constraint formulations $L_\theta$ that convert constraints into costs to be solved by an unconstrained optimizer, and adds their negative gradient to the update produced by SVGD for each particle. Importantly, the repulsive term in $\phi$ does not take into account the constraint, thus allowing particles to move together to satisfy constraints.

In the case of penalty or Lagrangian methods, the continuous approximation of the constraint pulls particles to the feasible region. 
If every particle is initialized in the feasible set, we can apply the same logic for the log barrier method, forcing particles to stay in the feasible set.

\begin{figure}[H]
    \centering
    \begin{subfigure}[b]{\linewidth}
            \centering
            \includesvg[width=\linewidth]{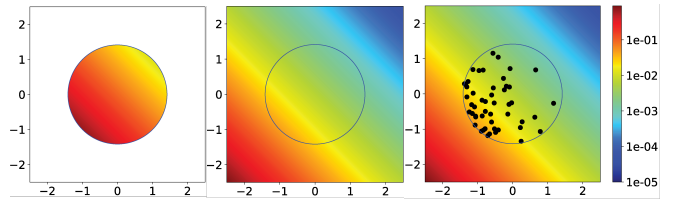}
            \caption{Desired distribution $p$ (left) vs approximation $\frac{1}{Z}e^{-\alpha f(x)}$ (center) vs distribution constructed with rejection sampling.  }
    \end{subfigure}
    \begin{subfigure}[b]{\linewidth}
        \centering
        Log Barrier\\
        \includesvg[width=\linewidth]{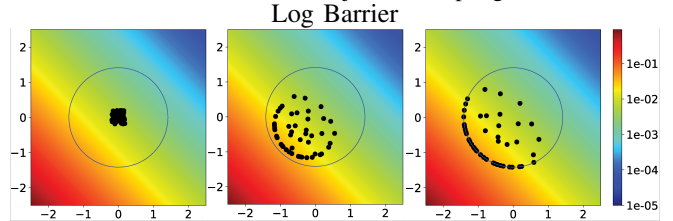}
        Augmented Lagrangian\\
        \includesvg[width=\linewidth]{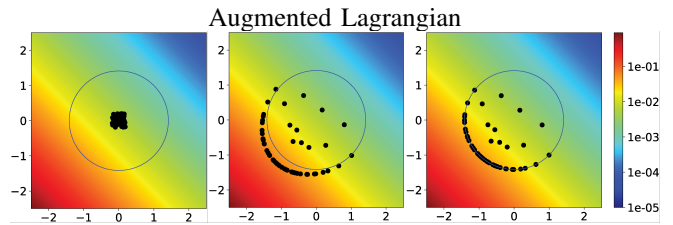}

        \caption{$Q$ method  with different constraint formulations over time. Shown at convergence of inner optimizer.}
    \end{subfigure}
    \begin{subfigure}[b]{\linewidth}
        \includesvg[width=\linewidth]{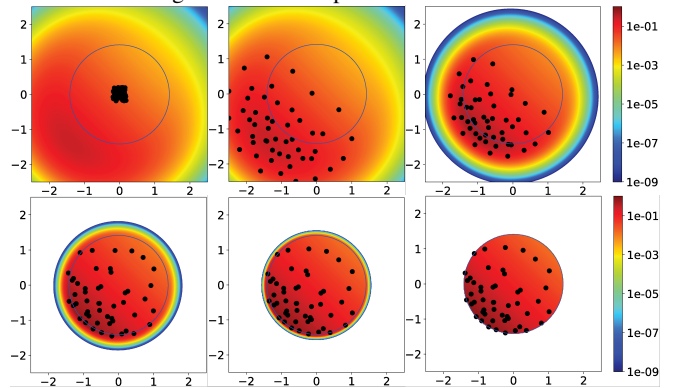}
         \caption{$\hat{p}$ method with augmented Lagrangian method as $d$ increases and Lagrange multipliers update each outer iteration. White regions have probability densities lower than the scale shown.}
    \end{subfigure}
    \caption{Our methods shown on $f(x) = x_1 + x_2$ with the inequality constraint $2-x_1^2 - x_2^2\geq 0$. To ensure it's finite, the total probability mass, $Z$, for each plot is only computed within the bound constraints.}
    \label{fig:method}
\end{figure}

After some amount of unconstrained optimizer steps, we can update the parameters of each of the constraints based on the particle that violates it the most or independently for each particle. 
When applicable, for faster convergence, we use backtracking line search to select the step length $\epsilon$. In \cite{projected_stein}, the authors proposed selecting the step length $\epsilon$ for each particle individually using backtracking line search with the objective function for particle $i$ at step $l$ as $-\log \left( p \left(z_j^l \right)\right)$. This selects step lengths for each particle that move it toward more probable regions. Similarly, we use $\sum_j^n \left[ -\log \left( p \left(z_j^l \right)\right) + L_\theta(z_j^l) \right]$ 
based on the summed negative log likelihood of every particle plus the constraint cost for each particle as our line search objective function. Using a shared $\epsilon$ for all particles lets us compute the objective function as a single batch operation and is more faithful to the particle update rule for SVGD. It is worth noting however that the repulsive term of SVGD is not represented in the objective function, so steps that exclusively increase diversity at the cost of lowering the likelihood of each of the particles will not be taken.

As an example, we show gradient descent for an equality constrained problem using the quadratic penalty method at iteration $k$ for particle $l$ in Eq.~\ref{eq:equalityUpdate}. 
\begin{equation}
    q_{k+1}^l = q_{k}^l + \epsilon_k \left(\!\!\!\!\!\!\!\!\!\!\!\begin{array}{cc}
         &      \frac{1}{n} \sum_j^n \left[ \alpha k(q_k^l,q_k^j)  \nabla_{q_k^l} f(q_k^l) - \nabla_{q_k^l} k(q_k^l,q_k^j) \right]\\
         & - 2 \mu h(q_k^l) \nabla h(q_k^l) 
    \end{array} \!\!\! \right)
    \label{eq:equalityUpdate}
\end{equation}


The approach in \cite{fabioLagrange} fits directly our paradigm, as an example of the $Q$ method, where $L_\theta$ was chosen to be the augmented Lagrangian function. They choose to use ADAM to select step lengths and independently add $\nabla L_\theta$ for each constraint to each particle.
On the other hand, CVSTO is similar to our $Q$ method, but they update particles with $\bm{q} = \bm{q} + \epsilon( \nabla_{\bm{q}} KL(\bm{q},p) - (\nabla_{\bm{q}} \nabla_{\bm{q}} L_\theta(\bm{q}))^{-1} \nabla_{\bm{q}} L_\theta(\bm{q}))$ \cite{power2023constrained}. 

In Fig.~\ref{fig:method}(a), we visualize the true distribution $p$, as defined in Eq~\ref{eq:true_dist}, the simple approximation $\frac{1}{Z}e^{-\alpha f(x)}$ and a ground truth distribution constructed with rejection sampling for a simple problem from \cite{NoceWrig06}. This problem has cost  $f(x) = x_1 + x_2 $ with the inequality constraint $2-x_1^2 - x_2^2 \geq 0$.  The global optimum of this optimization problem, and thus the MAP estimate of $p$ is $(-1,-1)$.

In Fig~\ref{fig:method}(b) we show $q$, containing 50 particles, trying to approximate $p$. We show the output of the inner optimizer using both an augmented Lagrangian and log barrier. We chose to use the RBF kernel with the median heuristic, as suggested by \cite{liu2019stein}. Similar to \cite{toussaint2024nlpsamplingcombiningmcmc} we compute the Earth Mover Distance between the constructed distribution and a ground truth distribution made with rejection sampling. Using distributions of 2000 particles, we find that the Augmented Lagrangian method results in the most accurate distribution. The EMD for augmented Lagrangian, log barrier, relaxed log barrier and quadratic penalty method is 0.155, 0.222, 0.231, 0.201 respectively.

\subsection{Modifying p}

The second formulation we propose is to iteratively modify the distribution $p$, that we are approximating, so that it accounts for the infeasible values being unlikely. The full $p$ method algorithm is shown in Alg.~\ref{algo:p}.

\begin{algorithm}
\caption{$p$  Constrained SVGD}
\label{algo:p}
\begin{algorithmic}[1]
    \State $q \gets k$ points sampled from distribution $D$
    \State $\hat{p} \gets  \frac{1}{Z} \left( e^{-\alpha f(m(x))} \prod e^{-L_\theta(m(x))}\right)$
    \While{not converged}
    \While{not converged}
    \State $\forall i \in k \quad \bm{q}_i \gets \bm{q}_i + \epsilon   \phi^*_{\hat{p}}(\bm{q}_i) $
    \EndWhile
    \State $\theta \gets \text{constraint\_param\_update}(\theta,\bm{q})$
    \State $\hat{p} \gets  \frac{1}{Z} \left( e^{-\alpha f(m(x))} \prod e^{-L_\theta(m(x))}\right)$
    \EndWhile
\end{algorithmic}
\end{algorithm}

Given a soft constraint formulation $L$ parameterized by $\theta$, we construct smooth approximations \\ $\hat{p} = \frac{1}{Z} \left( e^{-\alpha f(x)} \prod e^{-L_\theta(x)}\right)$ where infeasible points are less likely. Then we iteratively use SVGD to approximate $\hat{p}$. 
Note that with the $p$ method because $L$ is inside the distribution, the constraint formulation parameters $\theta$ are shared for all particles. 
This is similar to the methods that formulate constraints (e.g. collisions) into fixed cost, although here we update the weights~\cite{lambert2021stein,Dong2016MotionPA}.

Also, importantly, the gradient of $L$ enters into the SVGD equation, so particles use the kernel function to share constraint terms.
To handle box constraints under this paradigm, we could treat them as inequality constraints using our soft constraint formulation $L$. Alternatively, we can guarantee the box constraints by using a mapping function $m$ to map unbounded values of $x$ into $y$, where the range of $y=m(x)$ and thus $y$ is bounded (e.g. tanh/sigmoid/sin)~\cite{Deisenroth2010_1000019799}. In this case $\hat{p} = \frac{1}{Z} \left( e^{-\alpha f(m(x))} \prod e^{-L_\theta(m(x))}\right)$. Different choices for mapping function $m$ result in different behavior. The 
sin function allows $y$ to actually reach its extreme values, while the hyperbolic tangent and the sigmoid function only reach extremes asymptotically. The sin function however is not a unique mapping, so there are multiple equivalent regions of attraction.

As an example, we show the quadratic penalty method for equality constraints approximating $p$. 
\begin{equation}
    \hat{p}(x) = \frac{1}{Z} \left( e^{-\alpha f(x)} \prod_i e^{-\mu h_i(x)^2}\right)
\end{equation}

In Fig.~\ref{fig:method}(c) we visualize $\hat{p}$ as constructed by the augmented Lagrangian method as $d$ increases and Lagrange multipliers update each outer iteration. 
Different choices of $L$ result in different distributions $\hat{p}$, for example with the log barrier method, infeasible regions have zero probability. 

With the $p$ method, we are using the constraint terms to create a probability density shared across all particles. One interesting side effect of this is that Lagrange multipliers in the augmented Lagrangian method have to be shared for all particles. To update the Lagrange multipliers, we need a pooling operation to find which error to use per constraint. We choose to use the minimum value, the least in violation. This results in the solver defaulting to quadratic penalty method on constraints with some feasible particles. The other straightforward alternative, the maximum value, performed significantly worse in our limited testing. The most extreme particle in violation would create a large multiplier, and the other particles became unstable. Other possible pooling options like a median or mean pooling or more sophisticated multiplier update regime were not tested. 

To select steps sizes, we again recommend using backtracking line search with  $\sum_j^n -\log \left( \hat{p} \left(z_j^l \right)\right)$ as the objective function when possible. This selects step lengths that increase the total probability, under $\hat{p}$, of each of the particles without any regard for diversity. The diversity is still improved during the optimization, but when all particles settle in high quality solutions and the only gradient remaining is from the repulsion term, we select a small step size.

Similarly to with the $Q$ method we find with the $p$ method that the Augmented Lagrangian method results in the most accurate distribution according to the EMD. The EMD for augmented Lagrangian, log barrier, relaxed log barrier and quadratic penalty method is 0.089, 0.170, 0.134, 0.089 respectively. It's important to note that the $p$ method has substantially lower distance than the $Q$ method.

\section{Experiments}
To demonstrate our framework for constrained variational inference using Stein Variational gradient descent, we run experiments on three realistic robotics problems. First, we will identify a number of unique high quality motion plans that avoid obstacles by converting a trajectory optimization problem into a variational inference problem. Second, we will convert a robot arm inverse kinematics problem to approximate a Gaussian distribution in SE(3) with equality constraints that the robot wrist must be pointing upwards, and that the height exactly matches the desired height. These are desirable constraint for object placement or when holding objects that can spill. Lastly, we build a distribution over possible object states with Stein ICP while guaranteeing the object is vertical on top of a circular table.

\subsection{Trajectory Optimization, with inequality constraints}

\begin{figure}
    \begin{subfigure}[b]{\linewidth}
            \includegraphics[width=0.25\linewidth,trim={2.7cm 0.9cm 2.7cm 0.2cm},clip]{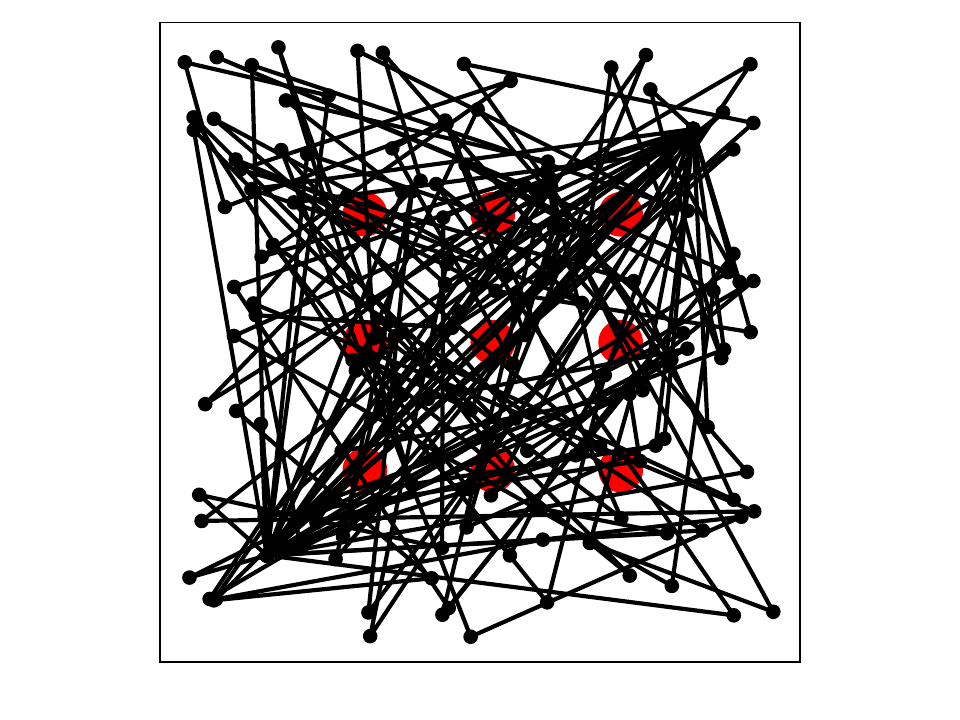}%
            \includegraphics[width=0.25\linewidth,trim={2.7cm 0.9cm 2.7cm 0.2cm},clip]{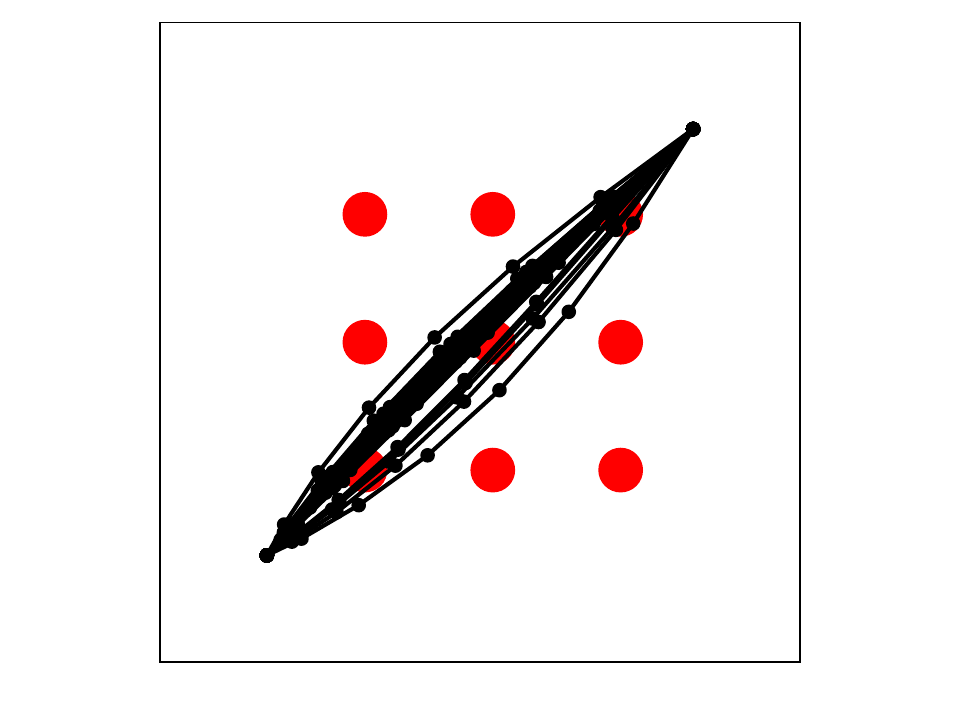}%
            \includegraphics[width=0.25\linewidth,trim={2.7cm 0.9cm 2.7cm 0.2cm},clip]{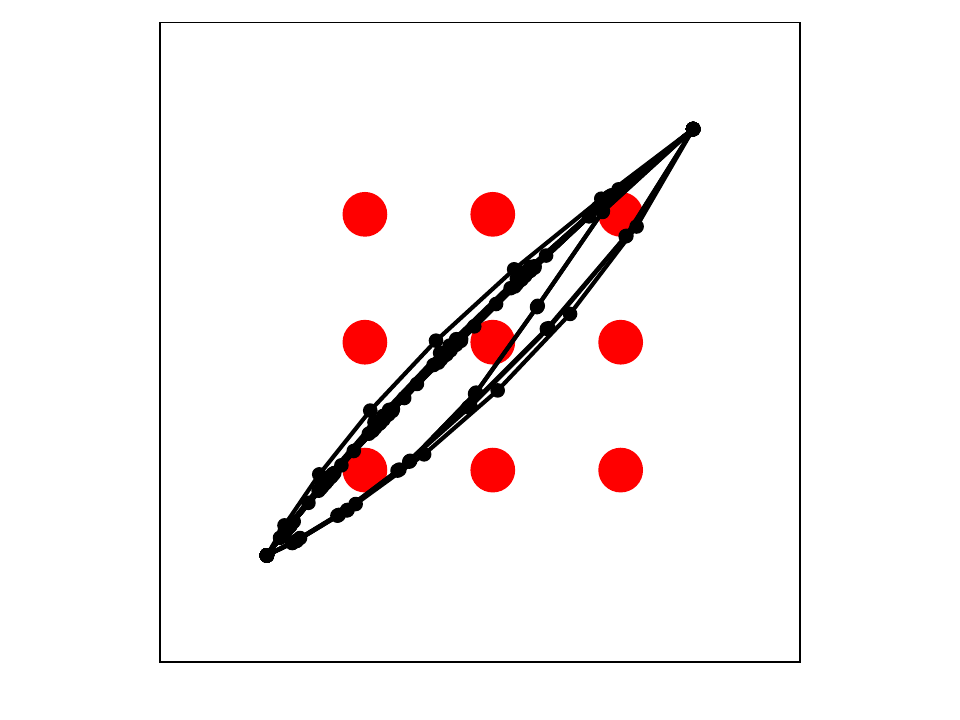}%
            \includegraphics[width=0.25\linewidth,trim={2.7cm 0.9cm 2.7cm 0.2cm},clip]{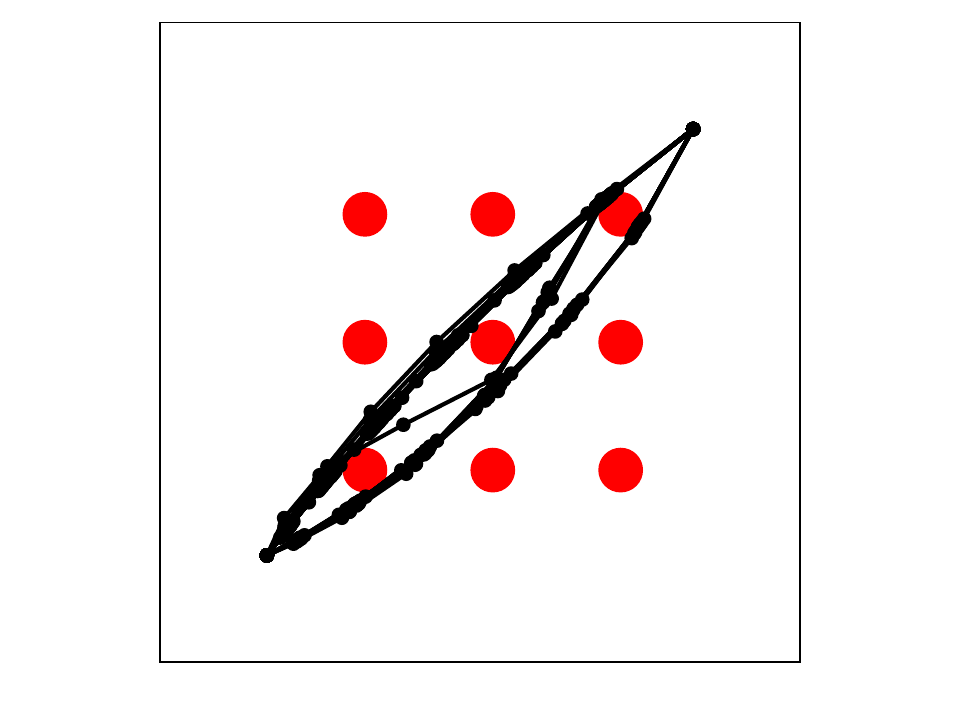}%
            \caption{Independent trajectory optimizations  }
    \end{subfigure}
    \begin{subfigure}[b]{\linewidth}
            \includegraphics[width=0.25\linewidth,trim={2.7cm 0.9cm 2.7cm 0.2cm},clip]{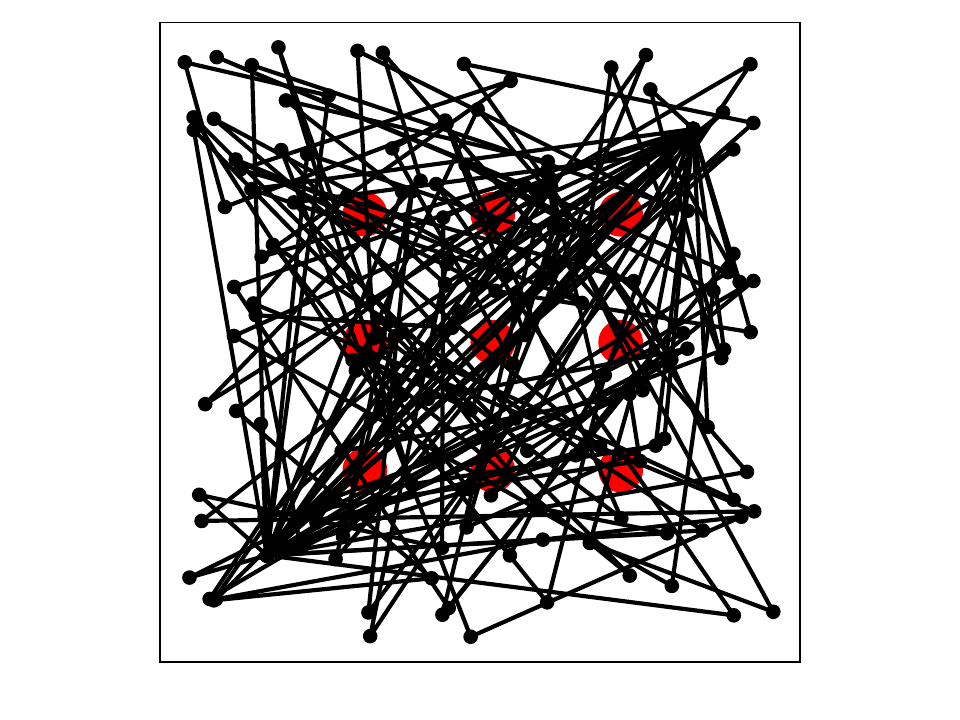}%
            \includegraphics[width=0.25\linewidth,trim={2.7cm 0.9cm 2.7cm 0.2cm},clip]{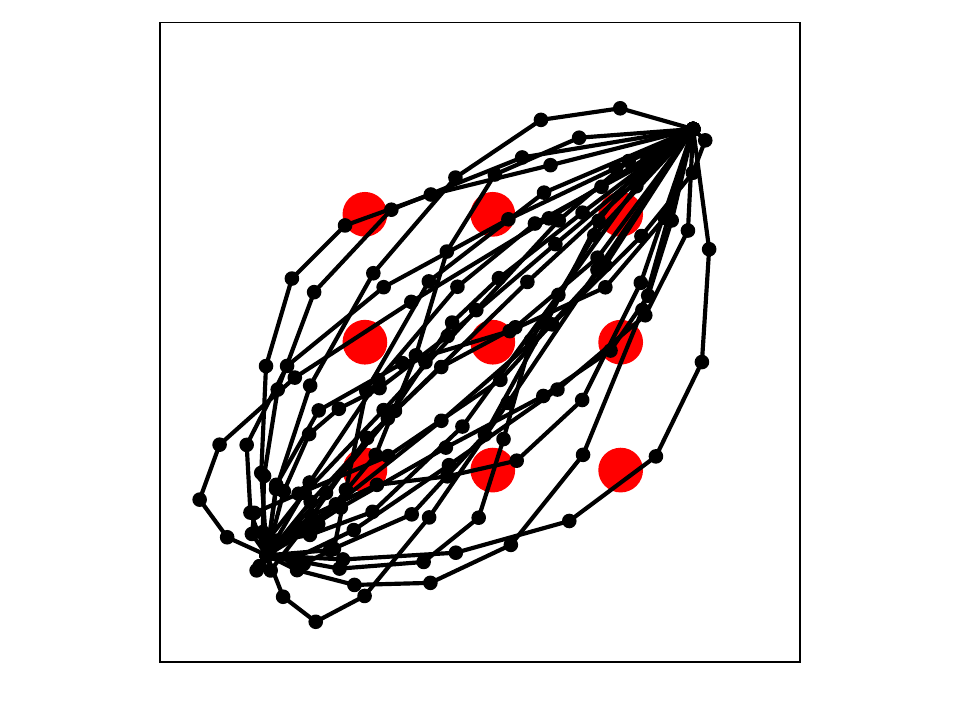}%
            \includegraphics[width=0.25\linewidth,trim={2.7cm 0.9cm 2.7cm 0.2cm},clip]{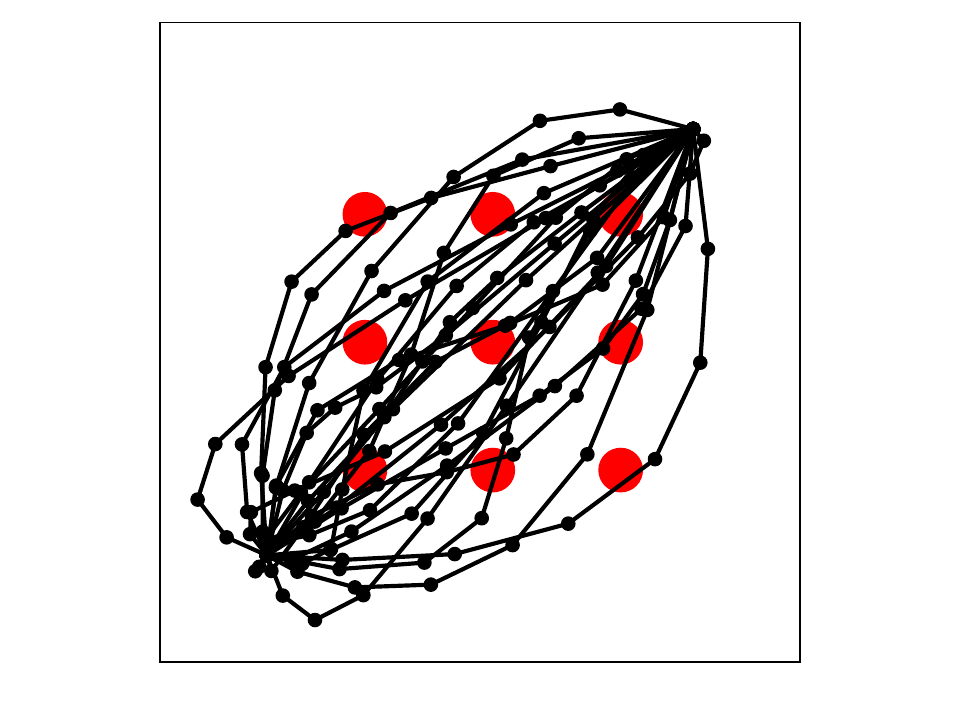}%
            \includegraphics[width=0.25\linewidth,trim={2.7cm 0.9cm 2.7cm 0.2cm},clip]{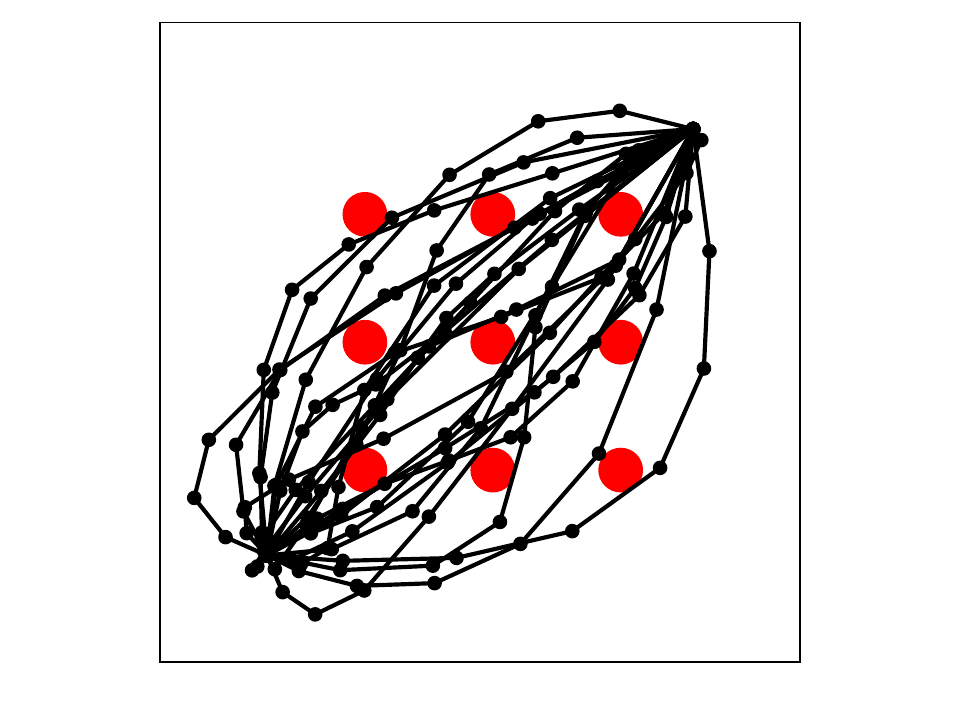}%
            \caption{$Q$ method }
    \end{subfigure}
    \begin{subfigure}[b]{\linewidth}
            \includegraphics[width=0.25\linewidth,trim={2.7cm 0.9cm 2.7cm 0.2cm},clip]{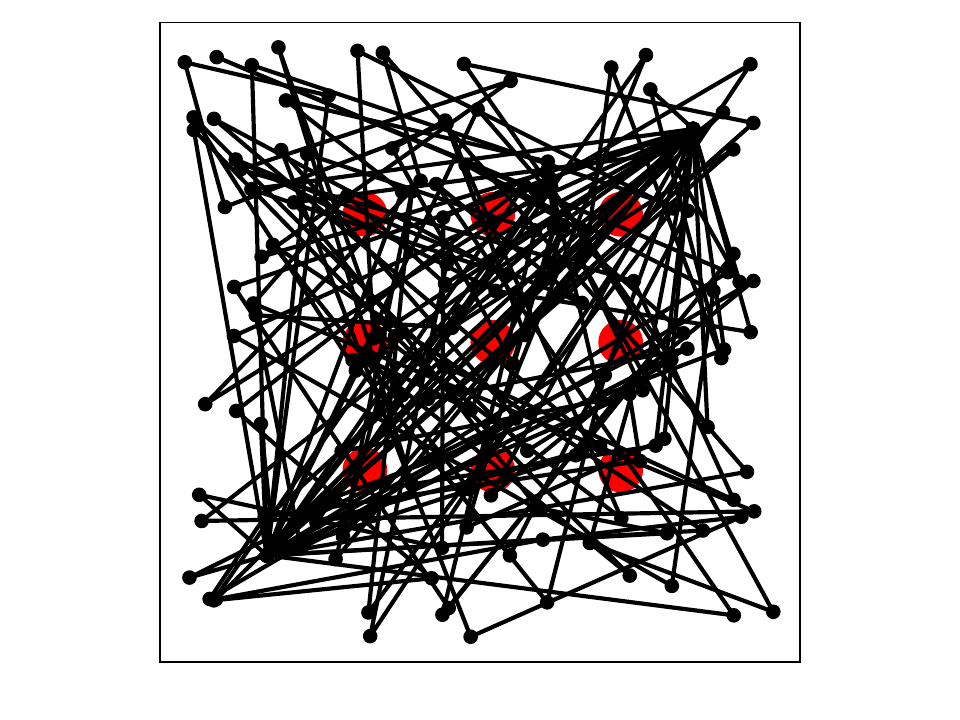}%
            \includegraphics[width=0.25\linewidth,trim={2.7cm 0.9cm 2.7cm 0.2cm},clip]{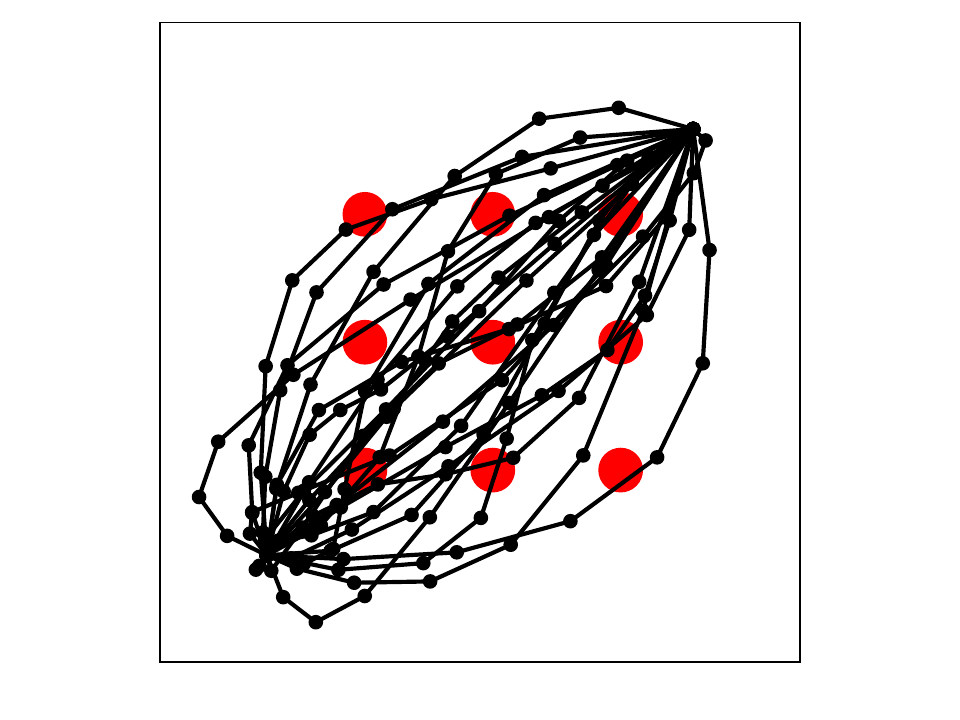}%
            \includegraphics[width=0.25\linewidth,trim={2.7cm 0.9cm 2.7cm 0.2cm},clip]{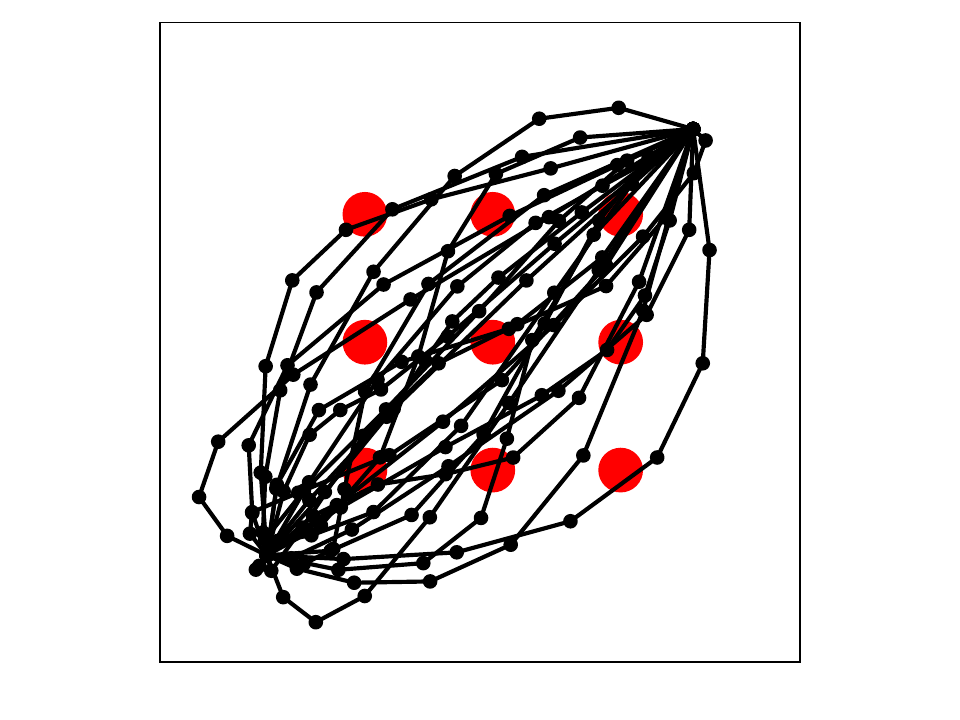}%
            \includegraphics[width=0.25\linewidth,trim={2.7cm 0.9cm 2.7cm 0.2cm},clip]{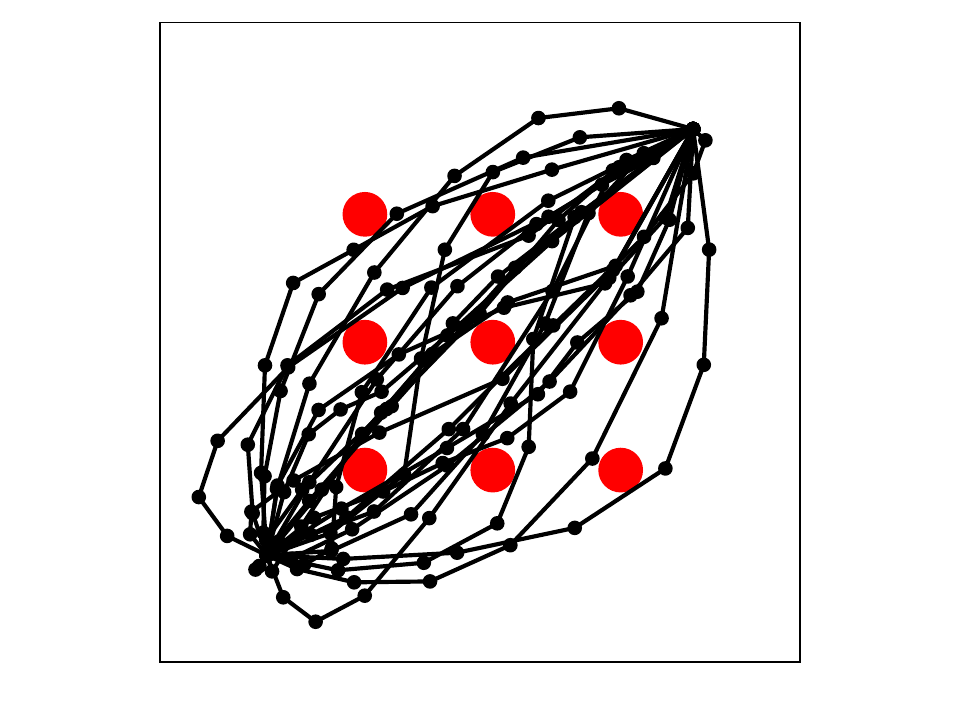}%
            \caption{$p$ method}
    \end{subfigure}
    \caption{Trajectory optimization problem initialized identically with our two different formulations and without Stein. Left to right shows outer optimization steps updating the constraint parameter terms after each inner optimizer converges.}
    \label{fig:traj}
\end{figure}
Similar to \cite{lambert2021stein} we would like to build a distribution of low cost 2D motion plans that avoid obstacles. However, we would like each plan to avoid obstacles to an arbitrary tolerance without being needlessly conservative.
For this task, we constrain the problem to start and end at desired points, and use 2D points along the trajectory as decision variables. The cost of a trajectory $f(x)$ is computed as the $\ell$2-norm of the derived acceleration between the waypoints. There are inequality constraints between the trajectory, the lines connecting waypoints, and the signed distance to each circular obstacle. 
We tune the hyperparameters by solving a single trajectory optimization problem with the augmented Lagrangian method and gradient descent. 
We believe this mimics the use case of practitioners already having existing implementations to find a single solution, and wanting to expand to multiple high quality solutions. We are able to show similar final results with both our formulations. We use the RBF kernel with median heuristic. 

Using the augmented Lagrangian, and running the inner loop until convergence, and show trajectories at each outer iteration in Fig~\ref{fig:traj}. Looking at these results, we can see how a single fixed penalty terms would produce trajectories with varying levels of constraint violation. By iterating on the constraint terms, we are able to find the smallest penalty term that still results in arbitrarily precise constraint satisfaction.

Note the results look very similar between $Q$ and $p$ methods as they were initialized the same and with a small value for $\theta$ (when $d=0,\gamma=0$ the methods both collapse to traditional SVGD). It's important to initialize with a small $d$ because it allows the trajectories to find low cost regions without being completely stuck in the homotopy class they were initialized. 
Alternatively, we compare to running independent trajectory optimization multiple times, which results in duplicate trajectories. We see much less diversity.   

\begin{figure}
    \centering
    \includegraphics[width=0.7\linewidth]{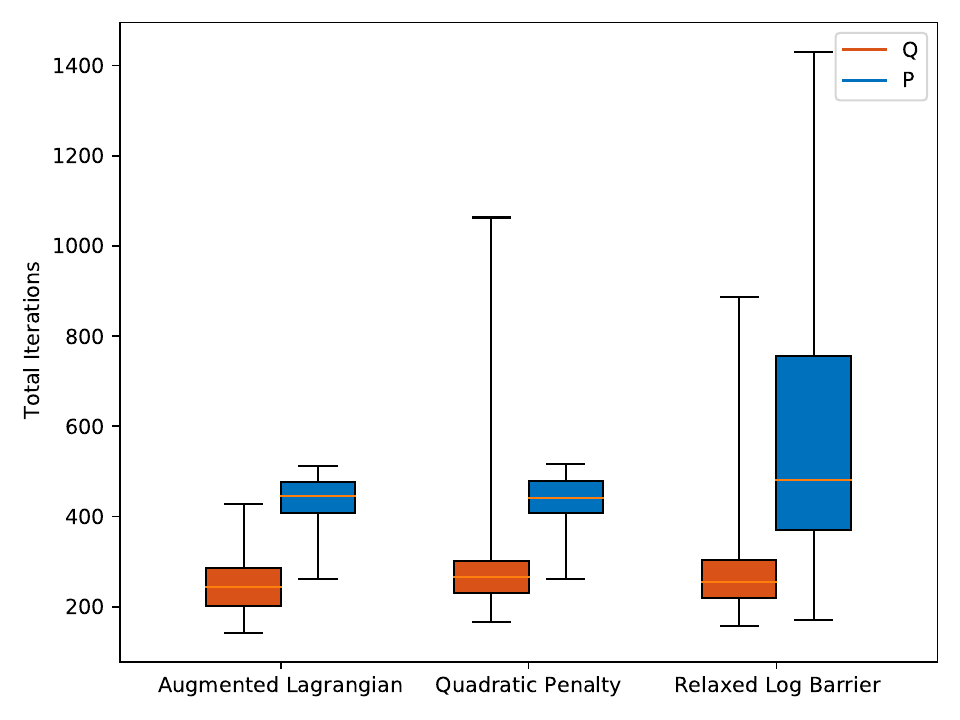}
    \caption{Total gradient steps until convergence under different permutations of our method over 50 trials each. }
    \label{fig:traj_iterations}
\end{figure}

In Fig.~\ref{fig:traj_iterations} we show the number of gradient steps required to converge using different permutations of our method. 
We see that the augmented Lagrangian method takes fewer iterations than the quadratic penalty method. Importantly, we also see that the $p$ method, where we share the constraint term through SVGD, consistently takes more iterations to converge than the $Q$ method.

We attribute this difference in performance to the following.
The inequality constraint creates a highly non-smooth gradient function. Infeasible trajectories share their large gradients with feasible trajectories pushing them into more conservative positions. Then, if those infeasible particles cross the constraint boundary and become feasible, the stable point changes and the original feasible point has to restabilize to a low-cost less conservative region.

\subsection{Inverse Kinematics over SE(3) Gaussian}\
\begin{figure}
    \centering
    \begin{subfigure}[b]{\linewidth}
        \centering
        \includegraphics[width=\linewidth,trim={0cm 0cm 0cm 0cm},clip]{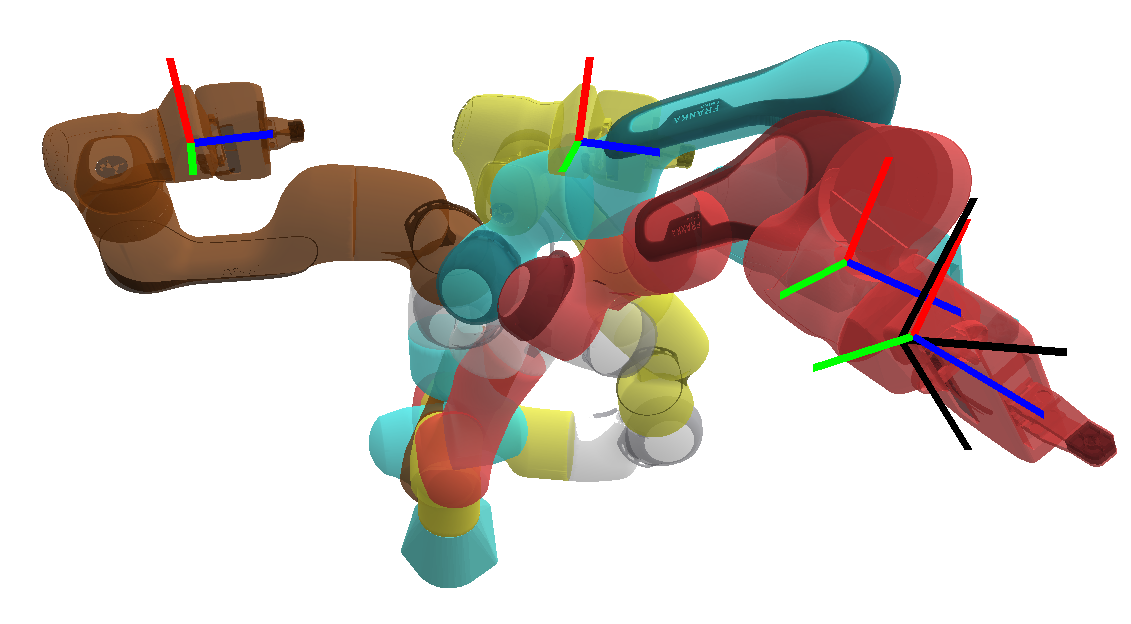}\\
        \includegraphics[width=0.7\linewidth,trim={0cm 0cm 0cm 0cm},clip]{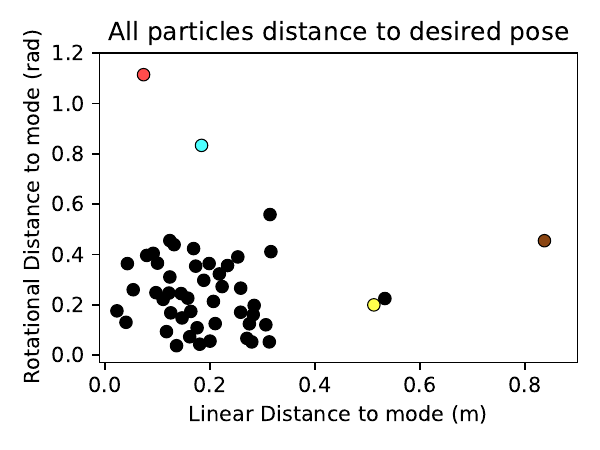}%
        \caption{50 particles We visualize the 4 solutions furthest from each other.}  
    \end{subfigure}
    \begin{subfigure}[b]{\linewidth}
        \centering
        \includegraphics[width=0.7\linewidth,trim={0cm 0cm 0cm 0cm},clip]{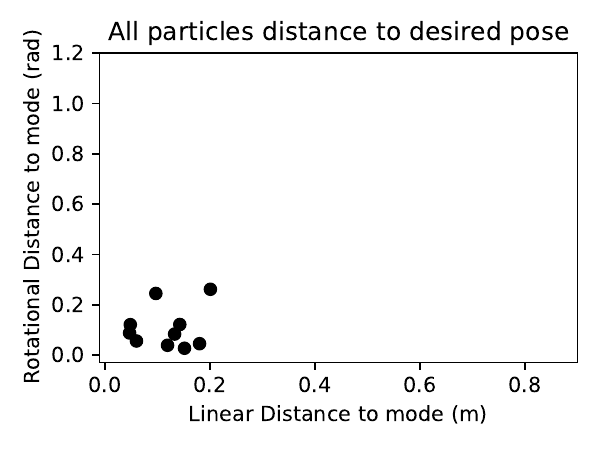}%
        \caption{10 particles}
    \end{subfigure}
    \begin{subfigure}[b]{\linewidth}
        \centering
        \includegraphics[width=0.7\linewidth,trim={0cm 0cm 0cm 0cm},clip]{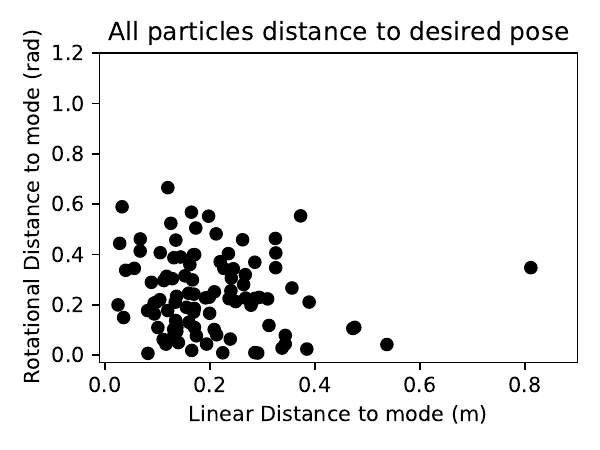}%
        \caption{100 particles}
    \end{subfigure}
    \caption{Constrained Stein variational gradient descent using to approximate Gaussian of end effector poses with equality constraints. All particles are shown by their rotational and linear distance to the target pose (mode). Note, while all particles have a rotational and linear distance from the mode pose, it is only along  axes without constraints. }
    \label{fig:IK}
\end{figure}

For our robot arm task, we take a standard robot arm inverse kinematics problem
$f(x) = ||\log(\texttt{fk}(x)^{-1} T)||_W^2$ 
where the function $\texttt{fk}$ computes the transformation matrix for robot arm joint angles using known forward kinematics and the $\log$ computes the screw associated with a transformation matrix. The resulting distribution $p(x)=e^{-\alpha f(x)}$ is now the Gaussian distribution on the SE(3) manifold. We have the following equality constraints $h_1(x) = ||\log(\texttt{fk}(x))_{3:5}||^2$ and $h_2(x) = ||\log(\texttt{fk}(x))_z||^2$ where $3:5$ corresponds to $x$ and $y$ rotational components of the screw and $z$ is the position error in the $z$ axis. These equality constraints are meant to be interpreted as placement/grasp constraints with respect to an object on a table. The robot arm also importantly has joint limits, that are encoded as bound constraints. We would like to build a distribution of joint angles that all are near a desired pose and obey our equality constraints.

If we solve this constrained problem independently 50 times, we see the vast majority of the initializations achieve the desired pose exactly. A few runs get stuck on bound constraints away from the desired pose, but still are able to find positions that obey the constraints.

Instead, we approximate this distribution with SVGD, using the RBF Kernel on the joint angles. In Fig.~\ref{fig:IK} we visualize the results with the $Q$ method using augmented Lagrangian cost function for a variety of different number of particles. We see that with more particles, the distribution spreads out farther from the desired pose (mode of $p$).  

However, for this equality constrained problem, we find that the $p$ method struggles. 
As the feasible set for our equality constraint is a lower dimensional manifold, sharing the penalty gradients results in particles not moving directly towards the feasible manifold.
With the $Q$ method as the penalty terms increase in magnitude they begin to dominate, and particles move directly towards the manifold and quickly become feasible. While with the $p$ method increasing the penalty multiplier doesn't necessarily converge to pointing towards the feasible set because other nearby particles share gradients that don't point directly towards the feasible manifold. Using 50 particles the $Q$ method, both quadratic penalty and augmented Lagrangian were able to converge in roughly 4000 total gradient steps, the $p$ method failed to converge in 25 times that many steps. For 10 particles, the $p$ method is able to converge after 100,000 iterations.


\subsection{State Estimation}
\begin{figure}
    \centering
    \includegraphics[width=\linewidth]{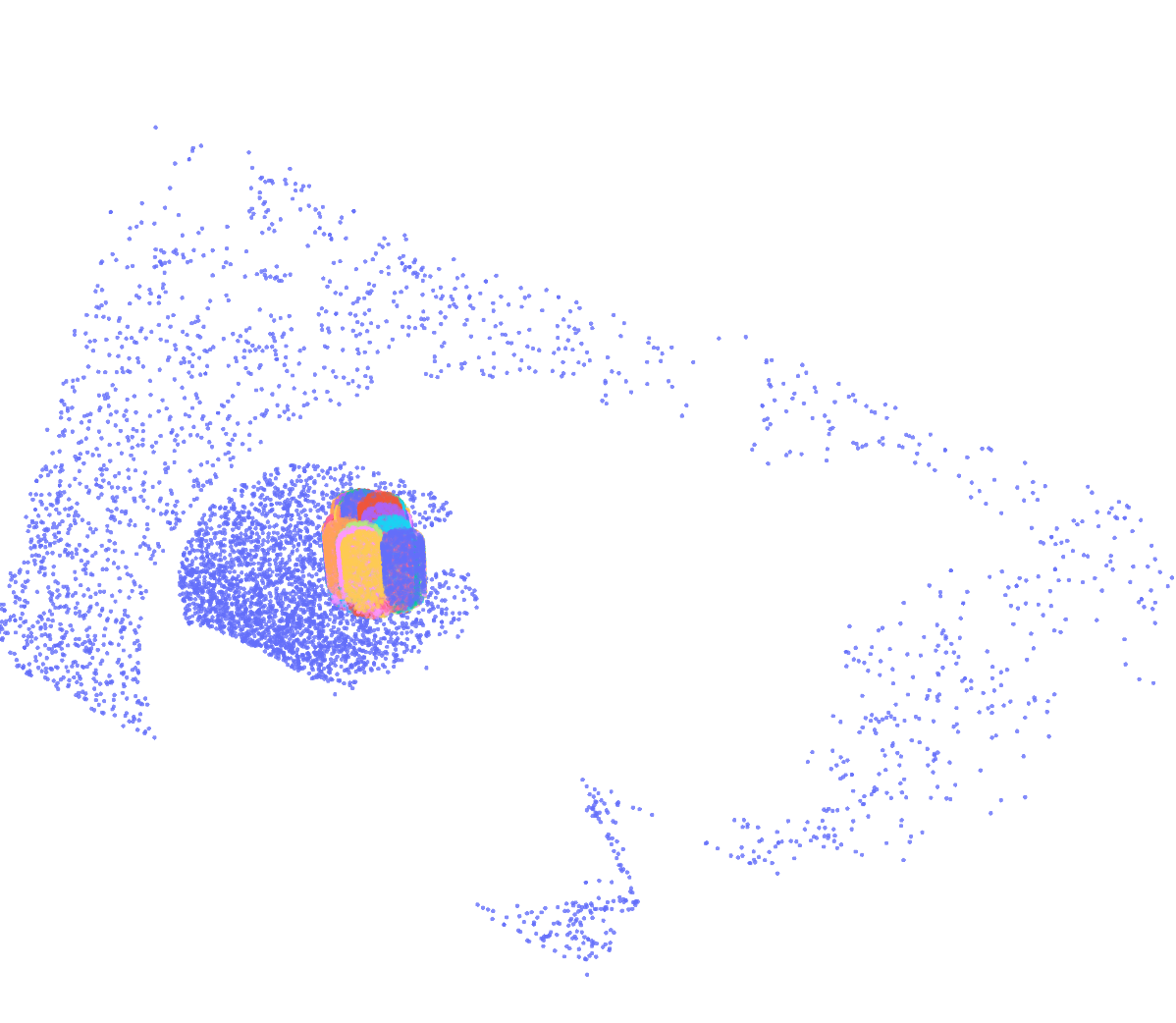}\\
    Univariate Kernel Density Estimate
    \includegraphics[width=0.49\linewidth]{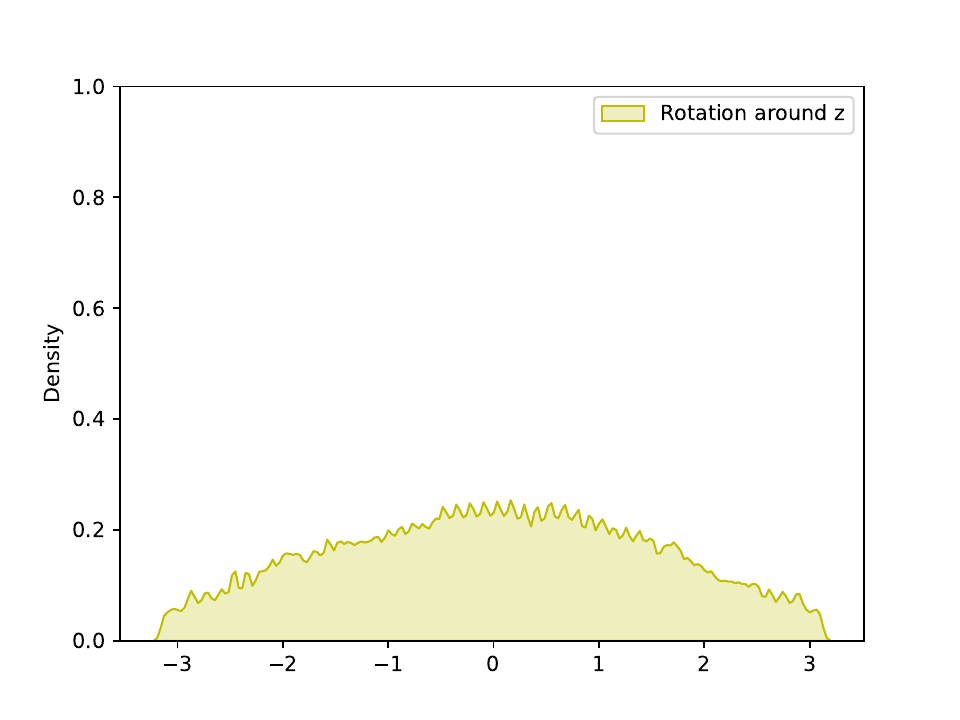} %
    \includegraphics[width=0.49\linewidth]{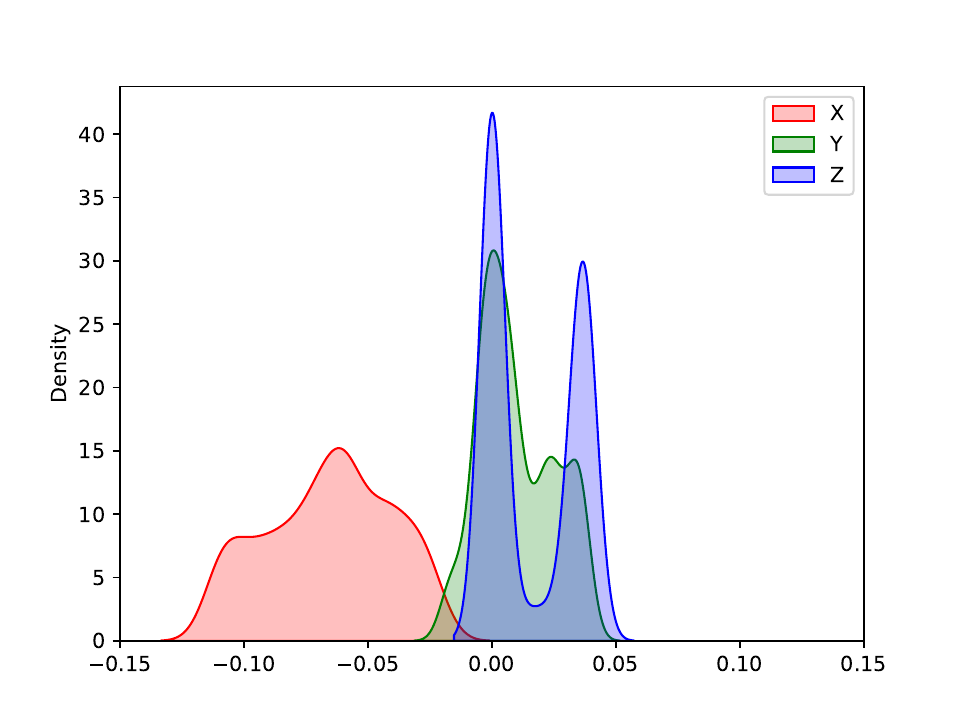}

    \caption{300 particles approximating the distribution created by the negative exponential of the ICP cost function. Partial view point cloud from RGBD camera, object occlusion shadow shown as empty stripe in blue circular region. }
    \label{fig:ICP_with_constraints}
\end{figure}

For our state estimation example, we examine a method based on Stein ICP. We define the ICP cost function as 
\begin{align}
    o :&= \{o_i \in O\}\\
    d(x) &= \underset{s\in S}{\min} ||s - x||_2 \\
    f(x) &= \frac{1}{1+N}\sum_i^N  
     \left\{ \begin{array}{ll}%
        d(o_i) &  d(o_i) < d_{\texttt{max}} \\
        0 & \texttt{otherwise}
        \end{array} \right.
\end{align}
where $o$ is a random subset of points from an object point cloud $O$ and  $S$ is a scene point cloud. Every iteration, we sample a new $o$ and thus compute new reference points. We add the additional constraints to our problem: $z>0$ (the object is above the table), $\sqrt{x^2 + y^2} < r$ (the object is within the boundary of a known circular table centered at $0,0$) and $||\log(T(x))_{3:5}||^2$ (the object $z$ axis is vertical). As opposed to our other problems where the additional constraints made the problem harder, these constraints are chosen to make the problem easier. Using the $Q$ method and a quadratic penalty, we are able to build a distribution of object poses. We show the results in Fig.~\ref{fig:ICP_with_constraints}.

\begin{figure}
    \centering
    \includegraphics[width=0.49\linewidth]{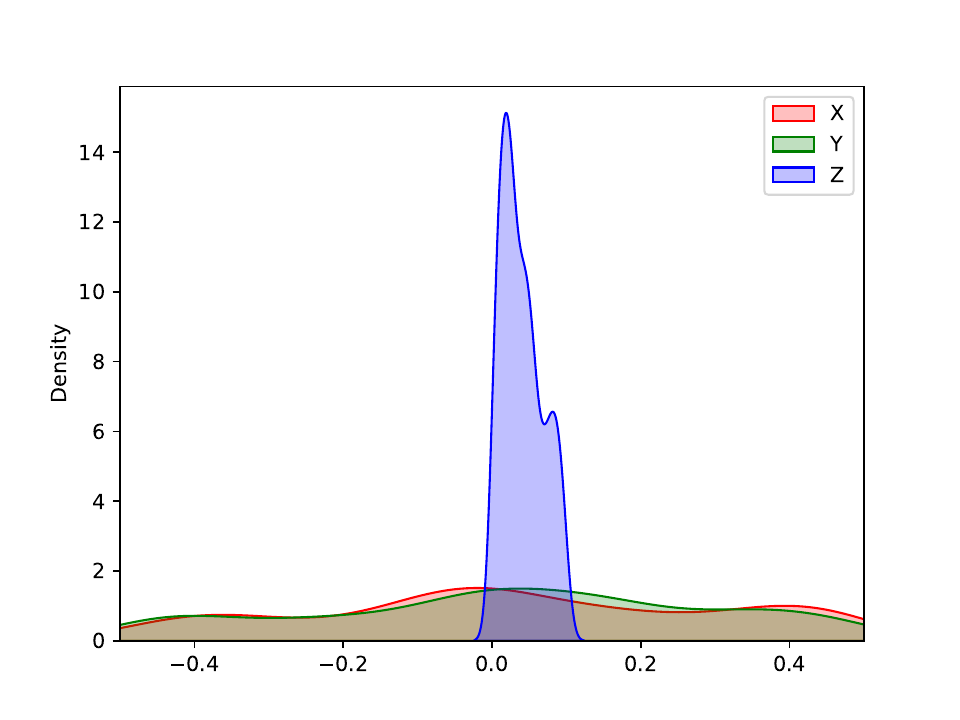}
    \caption{Identical problem to Fig.~\ref{fig:ICP_with_constraints} but without the constraints. }
    \label{fig:ICP_without_constraints}
\end{figure}
Due to the symmetry inherent to 3D rotations, we use $k(i,j) = \exp\frac{-||\log(T(i)^\intercal T(j))||_W^2}{h}$ with a hand tuned value $h$ based on the median heuristic. This directly compares the rotation matrices produced, rather than the difference in Euler angles as in Stein ICP.

While the original Stein ICP paper initializing their Euler angles uniformly from $[-0.043825,0.043825]$ for roll and pitch and $[-0.1753,0.1753]$ for yaw, we initialize our 3 vector of scaled axis angle values by uniformly sampling inside a sphere with radius $/pi$. In both cases, we sample positions uniformly from a broad set of positions.

We see from the kernel density computed on the particles that the $x$ and $y$ position is an unimodal distribution, the z position is bimodal. We expect the object size in the measured cloud is slightly different from the known cylinder, and the two modes are aligning the top vs the bottom of the object. We see we are able to recreate the results from the Stein ICP paper where symmetrical objects result in wide almost uniform distribution over rotation about $z$. As required by the constraint, the rotation about $y$ and $z$ are exactly zero for all particles.

We see in Fig.\ref{fig:ICP_with_constraints} compared to Fig.\ref{fig:ICP_without_constraints} that our additional constraints let us sample from a much wider range of orientations than Stein ICP and still converge to a distribution near the true object position.

\section{Conclusions}
We presented two broad formulations ($Q$ and $p$) for constrained Stein variational inference and how existing methods fit into our formulations. We further show that each formulation can be integrated in many distinct algorithmic forms. We explained how our approaches are derived as variational inference problems. We demonstrated that while both of our methods are able to effectively approximate distributions without violating constrains, the $Q$ method does so substantially faster on our test problems. The $Q$ method lets us efficiently achieve our task without molesting the underlying distribution. The $p$ method on the other hand struggles as the distribution it's approximating approaches a lower dimensional manifold.
It seems logical to expect problems exist where sharing constrain violation terms actually results in faster convergence but identifying those tasks is left as an open problem.

For our experiments here, we ran trials until convergence and until every particle was feasible. In cases where you desire unique high quality feasible solutions but don't need a full distribution, you can imagine using constrained Stein gradient descent as an anytime algorithm. If you initialize an arbitrarily large number of particles, over time the number of feasible particles will increase. By allocating more time, you can improve the spread of feasible solutions and have more options to choose between. 






\bibliographystyle{IEEEtran}

\bibliography{stein,model_learning}  

\begin{thebibliography}{10}
\providecommand{\url}[1]{#1}
\csname url@rmstyle\endcsname
\providecommand{\newblock}{\relax}
\providecommand{\bibinfo}[2]{#2}
\providecommand\BIBentrySTDinterwordspacing{\spaceskip=0pt\relax}
\providecommand\BIBentryALTinterwordstretchfactor{4}
\providecommand\BIBentryALTinterwordspacing{\spaceskip=\fontdimen2\font plus
\BIBentryALTinterwordstretchfactor\fontdimen3\font minus \fontdimen4\font\relax}
\providecommand\BIBforeignlanguage[2]{{%
\expandafter\ifx\csname l@#1\endcsname\relax
\typeout{** WARNING: IEEEtran.bst: No hyphenation pattern has been}%
\typeout{** loaded for the language `#1'. Using the pattern for}%
\typeout{** the default language instead.}%
\else
\language=\csname l@#1\endcsname
\fi
#2}}

\bibitem{prob-robotics}
S.~Thrun, W.~Burgard, and D.~Fox, \emph{Probabilistic robotics}, ser. Intelligent robotics and autonomous agents.\hskip 1em plus 0.5em minus 0.4em\relax {MIT} Press, 2005.

\bibitem{tabor2023adaptive}
\BIBentryALTinterwordspacing
G.~Tabor and T.~Hermans, ``Adaptive magnetic control using stein variational gradient descent computed distribution of object parameters,'' in \emph{IROS 2023 Workshop on Differentiable Probabilistic Robotics: Emerging Perspectives on Robot Learning}, 2023. [Online]. Available: \url{https://openreview.net/forum?id=dWE6dGTiTS}
\BIBentrySTDinterwordspacing

\bibitem{chance-constraint}
N.~E. Du~Toit and J.~W. Burdick, ``Robot motion planning in dynamic, uncertain environments,'' \emph{IEEE Transactions on Robotics}, vol.~28, no.~1, pp. 101--115, 2012.

\bibitem{Qingkai}
Q.~Lu, K.~Chenna, B.~Sundaralingam, and T.~Hermans, ``Planning multi-fingered grasps as probabilistic inference in a learned deep network,'' in \emph{Robotics Research}, N.~M. Amato, G.~Hager, S.~Thomas, and M.~Torres-Torriti, Eds.\hskip 1em plus 0.5em minus 0.4em\relax Cham: Springer International Publishing, 2020, pp. 455--472.

\bibitem{Thom-vi-mpc}
\BIBentryALTinterwordspacing
T.~Power and D.~Berenson, ``Variational inference {MPC} using normalizing flows and out-of-distribution projection,'' in \emph{Robotics: Science and Systems XVIII, New York City, NY, USA, June 27 - July 1, 2022}, K.~Hauser, D.~A. Shell, and S.~Huang, Eds., 2022. [Online]. Available: \url{https://doi.org/10.15607/RSS.2022.XVIII.027}
\BIBentrySTDinterwordspacing

\bibitem{vi_mp}
\BIBentryALTinterwordspacing
H.~Yu and Y.~Chen, ``A gaussian variational inference approach to motion planning,'' \emph{IEEE Robotics and Automation Letters}, vol.~8, no.~5, p. 2518–2525, May 2023. [Online]. Available: \url{http://dx.doi.org/10.1109/LRA.2023.3256134}
\BIBentrySTDinterwordspacing

\bibitem{toussaint2024nlpsamplingcombiningmcmc}
\BIBentryALTinterwordspacing
M.~Toussaint, C.~V. Braun, and J.~Ortiz-Haro, ``Nlp sampling: Combining mcmc and nlp methods for diverse constrained sampling,'' 2024. [Online]. Available: \url{https://arxiv.org/abs/2407.03035}
\BIBentrySTDinterwordspacing

\bibitem{Blei_2017}
\BIBentryALTinterwordspacing
D.~M. Blei, A.~Kucukelbir, and J.~D. McAuliffe, ``Variational inference: A review for statisticians,'' \emph{Journal of the American Statistical Association}, vol. 112, no. 518, pp. 859--877, apr 2017. [Online]. Available: \url{https://doi.org/10.1080%2F01621459.2017.1285773}
\BIBentrySTDinterwordspacing

\bibitem{kullback1951information}
S.~Kullback and R.~A. Leibler, ``On information and sufficiency,'' \emph{The annals of mathematical statistics}, vol.~22, no.~1, pp. 79--86, 1951.

\bibitem{yi2022sliced}
M.~Yi and S.~Liu, ``Sliced wasserstein variational inference,'' 2022.

\bibitem{campbell2019universal}
T.~Campbell and X.~Li, ``Universal boosting variational inference,'' 2019.

\bibitem{WangVIMPC}
\BIBentryALTinterwordspacing
Z.~Wang, O.~So, J.~Gibson, B.~I. Vlahov, M.~Gandhi, G.~Liu, and E.~A. Theodorou, ``Variational inference {MPC} using tsallis divergence,'' in \emph{Robotics: Science and Systems XVII, Virtual Event, July 12-16, 2021}, D.~A. Shell, M.~Toussaint, and M.~A. Hsieh, Eds., 2021. [Online]. Available: \url{https://doi.org/10.15607/RSS.2021.XVII.073}
\BIBentrySTDinterwordspacing

\bibitem{NEURIPS2020_c928d86f}
\BIBentryALTinterwordspacing
N.~Wan, D.~Li, and N.~HOVAKIMYAN, ``f-divergence variational inference,'' in \emph{Advances in Neural Information Processing Systems}, H.~Larochelle, M.~Ranzato, R.~Hadsell, M.~Balcan, and H.~Lin, Eds., vol.~33.\hskip 1em plus 0.5em minus 0.4em\relax Curran Associates, Inc., 2020, pp. 17\,370--17\,379. [Online]. Available: \url{https://proceedings.neurips.cc/paper_files/paper/2020/file/c928d86ff00aeb89a39bd4a80e652a38-Paper.pdf}
\BIBentrySTDinterwordspacing

\bibitem{MCMCCHIB20013569}
\BIBentryALTinterwordspacing
S.~Chib, ``Chapter 57 - markov chain monte carlo methods: Computation and inference,'' ser. Handbook of Econometrics, J.~J. Heckman and E.~Leamer, Eds.\hskip 1em plus 0.5em minus 0.4em\relax Elsevier, 2001, vol.~5, pp. 3569--3649. [Online]. Available: \url{https://www.sciencedirect.com/science/article/pii/S1573441201050103}
\BIBentrySTDinterwordspacing

\bibitem{MCMC_CVPR}
S.-C. Zhu, R.~Zhang, and Z.~Tu, ``Integrating bottom-up/top-down for object recognition by data driven markov chain monte carlo,'' in \emph{Proceedings IEEE Conference on Computer Vision and Pattern Recognition. CVPR 2000 (Cat. No.PR00662)}, vol.~1, 2000, pp. 738--745 vol.1.

\bibitem{barcelos2021dual}
L.~Barcelos, A.~Lambert, R.~Oliveira, P.~Borges, B.~Boots, and F.~Ramos, ``Dual online stein variational inference for control and dynamics,'' 2021.

\bibitem{lambert2021stein}
A.~Lambert, A.~Fishman, D.~Fox, B.~Boots, and F.~Ramos, ``Stein variational model predictive control,'' 2021.

\bibitem{honda2023stein}
K.~Honda, N.~Akai, K.~Suzuki, M.~Aoki, H.~Hosogaya, H.~Okuda, and T.~Suzuki, ``Stein variational guided model predictive path integral control: Proposal and experiments with fast maneuvering vehicles,'' 2023.

\bibitem{power2023constrained}
T.~Power and D.~Berenson, ``Constrained stein variational trajectory optimization,'' 2023.

\bibitem{sharma2023taskspace}
\BIBentryALTinterwordspacing
M.~S. Sharma, T.~Power, and D.~Berenson, ``Task-space kernels for diverse stein variational {MPC},'' in \emph{IROS 2023 Workshop on Differentiable Probabilistic Robotics: Emerging Perspectives on Robot Learning}, 2023. [Online]. Available: \url{https://openreview.net/forum?id=IeG4J5TF79}
\BIBentrySTDinterwordspacing

\bibitem{lee2023stamp}
Y.~Lee, P.~Huang, K.~M. Jatavallabhula, A.~Z. Li, F.~Damken, E.~Heiden, K.~Smith, D.~Nowrouzezahrai, F.~Ramos, and F.~Shkurti, ``Stamp: Differentiable task and motion planning via stein variational gradient descent,'' 2023.

\bibitem{stein-ICP-2106-03287}
\BIBentryALTinterwordspacing
F.~A. Maken, F.~Ramos, and L.~Ott, ``Stein {ICP} for uncertainty estimation in point cloud matching,'' \emph{CoRR}, vol. abs/2106.03287, 2021. [Online]. Available: \url{https://arxiv.org/abs/2106.03287}
\BIBentrySTDinterwordspacing

\bibitem{fan2021stein}
J.~Fan, A.~Taghvaei, and Y.~Chen, ``Stein particle filtering,'' 2021.

\bibitem{stein_multi_robot}
J.~Pavlasek, J.~J.~Z. Mah, R.~Xu, O.~C. Jenkins, and F.~Ramos, ``Stein variational belief propagation for multi-robot coordination,'' \emph{IEEE Robotics and Automation Letters}, vol.~9, no.~5, pp. 4194--4201, 2024.

\bibitem{altro}
T.~A. Howell, B.~E. Jackson, and Z.~Manchester, ``Altro: A fast solver for constrained trajectory optimization,'' in \emph{2019 IEEE/RSJ International Conference on Intelligent Robots and Systems (IROS)}, 2019, pp. 7674--7679.

\bibitem{Schulman2013FindingLO}
\BIBentryALTinterwordspacing
J.~Schulman, J.~Ho, A.~X. Lee, I.~Awwal, H.~Bradlow, and P.~Abbeel, ``Finding locally optimal, collision-free trajectories with sequential convex optimization,'' \emph{Robotics: Science and Systems IX}, 2013. [Online]. Available: \url{https://api.semanticscholar.org/CorpusID:2393365}
\BIBentrySTDinterwordspacing

\bibitem{bala_dynamics}
\BIBentryALTinterwordspacing
B.~Sundaralingam and T.~Hermans, ``In-hand object-dynamics inference using tactile fingertips,'' \emph{CoRR}, vol. abs/2003.13165, 2020. [Online]. Available: \url{https://arxiv.org/abs/2003.13165}
\BIBentrySTDinterwordspacing

\bibitem{inertia_consistent}
P.~M. Wensing, S.~Kim, and J.-J.~E. Slotine, ``Linear matrix inequalities for physically consistent inertial parameter identification: A statistical perspective on the mass distribution,'' \emph{IEEE Robotics and Automation Letters}, vol.~3, no.~1, pp. 60--67, 2018.

\bibitem{ICP_bounded_angle}
\BIBentryALTinterwordspacing
C.~Zhang, S.~Du, J.~Liu, Y.~Li, J.~Xue, and Y.~Liu, ``Robust iterative closest point algorithm with bounded rotation angle for 2d registration,'' \emph{Neurocomputing}, vol. 195, pp. 172--180, 2016, learning for Medical Imaging. [Online]. Available: \url{https://www.sciencedirect.com/science/article/pii/S0925231216001211}
\BIBentrySTDinterwordspacing

\bibitem{ICP_Kinematic_constraints}
\BIBentryALTinterwordspacing
T.~Guadagnino, B.~Mersch, I.~Vizzo, S.~Gupta, M.~V.~R. Malladi, L.~Lobefaro, G.~Doisy, and C.~Stachniss, ``Kinematic-icp: Enhancing lidar odometry with kinematic constraints for wheeled mobile robots moving on planar surfaces,'' 2024. [Online]. Available: \url{https://arxiv.org/abs/2410.10277}
\BIBentrySTDinterwordspacing

\bibitem{fabioLagrange}
\BIBentryALTinterwordspacing
E.~Heiden, C.~E. Denniston, D.~Millard, F.~Ramos, and G.~S. Sukhatme, ``Probabilistic inference of simulation parameters via parallel differentiable simulation,'' \emph{CoRR}, vol. abs/2109.08815, 2021. [Online]. Available: \url{https://arxiv.org/abs/2109.08815}
\BIBentrySTDinterwordspacing

\bibitem{zhang2022sampling}
R.~Zhang, Q.~Liu, and X.~T. Tong, ``Sampling in constrained domains with orthogonal-space variational gradient descent,'' 2022.

\bibitem{efk_ICP}
\BIBentryALTinterwordspacing
M.~Barczyk, S.~Bonnabel, J.~Deschaud, and F.~Goulette, ``Invariant {EKF} design for scan matching-aided localization,'' \emph{CoRR}, vol. abs/1503.01407, 2015. [Online]. Available: \url{http://arxiv.org/abs/1503.01407}
\BIBentrySTDinterwordspacing

\bibitem{stein-particle-filter}
F.~A. Maken, F.~Ramos, and L.~Ott, ``Stein particle filter for nonlinear, non-gaussian state estimation,'' \emph{IEEE Robotics and Automation Letters}, vol.~7, no.~2, pp. 5421--5428, 2022.

\bibitem{DBLP:journals/corr/abs-2103-12890}
\BIBentryALTinterwordspacing
L.~Barcelos, A.~Lambert, R.~Oliveira, P.~Borges, B.~Boots, and F.~Ramos, ``Dual online stein variational inference for control and dynamics,'' \emph{CoRR}, vol. abs/2103.12890, 2021. [Online]. Available: \url{https://arxiv.org/abs/2103.12890}
\BIBentrySTDinterwordspacing

\bibitem{kf_state_parameters}
M.~H. Riva, M.~Dagen, and T.~Ortmaier, ``Adaptive unscented kalman filter for online state, parameter, and process covariance estimation,'' in \emph{2016 American Control Conference (ACC)}, 2016, pp. 4513--4519.

\bibitem{yu2024stochasticmotionplanninggaussian}
\BIBentryALTinterwordspacing
H.~Yu and Y.~Chen, ``Stochastic motion planning as gaussian variational inference: Theory and algorithms,'' 2024. [Online]. Available: \url{https://arxiv.org/abs/2308.14985}
\BIBentrySTDinterwordspacing

\bibitem{liu2019stein}
Q.~Liu and D.~Wang, ``Stein variational gradient descent: A general purpose bayesian inference algorithm,'' 2019.

\bibitem{liu2016kernelized}
Q.~Liu, J.~D. Lee, and M.~I. Jordan, ``A kernelized stein discrepancy for goodness-of-fit tests and model evaluation,'' 2016.

\bibitem{NoceWrig06}
J.~Nocedal and S.~J. Wright, \emph{Numerical Optimization}, 2nd~ed.\hskip 1em plus 0.5em minus 0.4em\relax New York, NY, USA: Springer, 2006.

\bibitem{toussaint2014novel}
M.~Toussaint, ``A novel augmented lagrangian approach for inequalities and convergent any-time non-central updates,'' 2014.

\bibitem{relaxedlogbarrier}
R.~Grandia, F.~Farshidian, R.~Ranftl, and M.~Hutter, ``Feedback mpc for torque-controlled legged robots,'' in \emph{2019 IEEE/RSJ International Conference on Intelligent Robots and Systems (IROS)}, 2019, pp. 4730--4737.

\bibitem{relaxedlogbarrier_original}
J.~Hauser and A.~Saccon, ``A barrier function method for the optimization of trajectory functionals with constraints,'' in \emph{Proceedings of the 45th IEEE Conference on Decision and Control}, 2006, pp. 864--869.

\bibitem{projected_stein}
P.~Chen and O.~Ghattas, ``Projected stein variational gradient descent,'' in \emph{Proceedings of the 34th International Conference on Neural Information Processing Systems}, ser. NIPS '20.\hskip 1em plus 0.5em minus 0.4em\relax Red Hook, NY, USA: Curran Associates Inc., 2020.

\bibitem{Dong2016MotionPA}
\BIBentryALTinterwordspacing
J.~Dong, M.~Mukadam, F.~Dellaert, and B.~Boots, ``Motion planning as probabilistic inference using gaussian processes and factor graphs,'' in \emph{Robotics: Science and Systems}, 2016. [Online]. Available: \url{https://api.semanticscholar.org/CorpusID:17831189}
\BIBentrySTDinterwordspacing

\bibitem{Deisenroth2010_1000019799}
M.~P. Deisenroth, ``\BIBforeignlanguage{english}{Efficient reinforcement learning using gaussian processes},'' Ph.D. dissertation, 2010.

\end{thebibliography}

\end{document}